\documentclass[letterpaper, 10 pt, journal, twoside]{ieeetran}% Use this command for final RAL version

\IEEEoverridecommandlockouts                              % This command is only needed if 
\usepackage{cite}
\usepackage{amsmath,amssymb,amsfonts}
\usepackage{algorithmic}
\usepackage{graphicx}
\usepackage{textcomp}
\usepackage{gensymb}
\usepackage{xcolor}
\usepackage{color}		
\usepackage{wasysym}
\usepackage{multirow}		
\usepackage{array}										
\usepackage{colortbl}
\usepackage{siunitx}
\usepackage[english]{babel}	
\usepackage[hidelinks]{hyperref}     
\usepackage{blindtext}
\usepackage[ruled, linesnumbered]{algorithm2e}
\usepackage{pifont}
\title{
SPONGE: Open-Source Designs of Modular Articulated Soft Robots
}

\author{Tim-Lukas Habich$^{1*}$, Jonas Haack$^{1*}$, Mehdi Belhadj$^{1}$, Dustin Lehmann$^{2}$, Thomas Seel$^{1}$, and Moritz Schappler$^{1}$%
\thanks{Manuscript received November 21, 2023; revised February 15, 2024; accepted March 29, 2024.}%Use only for final RAL version
\thanks{This paper was recommended for publication by editor Y.-L. Park upon evaluation of the associate editor and reviewers' comments.
	This work was supported by the Deutsche Forschungsgemeinschaft (DFG, German Research Foundation) under grant no. 433586601.} %Use only for final RAL version
\thanks{$^{1}$The authors are with the Leibniz University Hannover, Institute of Mechatronic Systems, 30823 Garbsen, Germany,
{\tt\footnotesize{\href{mailto:tim-lukas.habich@imes.uni-hannover.de}{tim-lukas.habich@imes.uni-hannover.de}}}}%
\thanks{$^{2}$The author is with the Technical University of Berlin, Control Systems Group, \mbox{10587 Berlin, Germany.}}%
\thanks{$^{*}$Both authors contributed equally to this publication.}%
\thanks{$^{\dagger}$\tt\footnotesize{\url{https://tlhabich.github.io/sponge}}}%
\thanks{Digital Object Identifier (DOI): 10.1109/LRA.2024.3388855}
}

\newif\ifcopyright
	\copyrighttrue
\newif\ifhighlightchanges
\ifhighlightchanges
\newcommand{\highlightredsec}[1]{\textcolor{red}{#1}}
\else
\newcommand{\highlightredsec}[1]{#1}
\fi
\newcommand{\highlightred}[1]{#1}
\newcommand{\quot}[1]{``#1"}

\newcommand{\cmark}{\textcolor{green}{\ding{51}}}
\newcommand{\xmark}{\textcolor{red}{\ding{55}}}
\newcommand{\mm}[1]{\boldsymbol{#1}}

\newcommand{\ind}[1]{\mathrm{#1}}
\usepackage[mathscr]{euscript}
\newcommand{\R}{\mathbb{R}}

\newcommand{\transpose}{^\mathrm{T}}
\definecolor{Gray}{gray}{0.85}

\newcolumntype{M}[1]{>{\centering\arraybackslash}m{#1}}
\newcolumntype{N}{@{}m{0pt}@{}}
\makeatletter
\newcommand{\removelatexerror}{\let\@latex@error\@gobble}
\makeatother

\begin{document}
\ifcopyright
{\LARGE IEEE Copyright Notice}
\newline
\fboxrule=0.4pt \fboxsep=3pt

\fbox{\begin{minipage}{1.1\linewidth}  % <-- hier Kastenbreiter der Kopfzeile ändern
		%\centering
		% <-- hier Namen der Konferenz und Jahr einfügen
		%Changes were made to this version by the publisher prior to publication. \\
		%	The final version of record is available at http://doi.org/10.1109/XYZ123456789.TODO   % <-- hier DOI einfügen
		\textcopyright\,\,2024\,\,IEEE. Personal use of this material is permitted. Permission from IEEE must be obtained for all other uses, in any current or future media, including reprinting/republishing this material for advertising or promotional purposes, creating new collective works, for resale or redistribution to servers or lists, or reuse of any copyrighted component of this work in other works. \\
		
		Accepted to be published in: IEEE Robotics and Automation Letters (RA-L), 2024.\\
		
		DOI: 10.1109/LRA.2024.3388855
		
\end{minipage}}
\else
\fi
\graphicspath{{./images/}}

\maketitle
% Comment or remove these lines for final RAL version.
%\thispagestyle{empty}
%\pagestyle{empty}

\markboth{IEEE Robotics and Automation Letters. Preprint Version. Accepted March, 2024}
{Habich \MakeLowercase{\textit{et al.}}: Open-Source Designs of Modular Articulated Soft Robots}

%%%%%%%%%%%%%%%%%%%%%%%%%%%%%%%%%%%%%%%%%%%%%%%%%%%%%%%%%%%%%%%%%%%%%%%%%%%%%%%%
\begin{abstract}
	Soft-robot designs are manifold, but only a few are publicly available. Often, these are only briefly described in their publications. This complicates reproduction, and hinders the reproducibility and comparability of research results. If the designs were uniform and open source, validating researched methods on real benchmark systems would be possible. \highlightred{To address this, we present two variants of a soft pneumatic robot with antagonistic bellows as open source.} Starting from a semi-modular design with multiple cables and tubes routed through the robot body, the transition to a fully modular robot with integrated microvalves and serial communication is highlighted. Modularity in terms of stackability, actuation, and communication is achieved, which is the crucial requirement for building soft robots with many degrees of freedom and high dexterity for real-world tasks. \highlightred{Both systems are compared regarding their respective advantages and disadvantages.} The robots' functionality is demonstrated in experiments on airtightness, \highlightredsec{gravitational influence,} position control with mean tracking errors of ${<}\SI{3}{deg}$, and long-term operation of cast and printed bellows. \highlightred{All soft- and hardware files required for reproduction are provided$^\dagger$.}
	
\end{abstract}

%%%%%%%%%%%%%%%%%%%%%%%%%%%%%%%%%%%%%%%%%%%%%%%%%%%%%%%%%%%%%%%%%%%%%%%%%%%%%%%%

\section{Introduction}
\begin{figure}[t]
		\centering
		\resizebox{1\linewidth}{!}{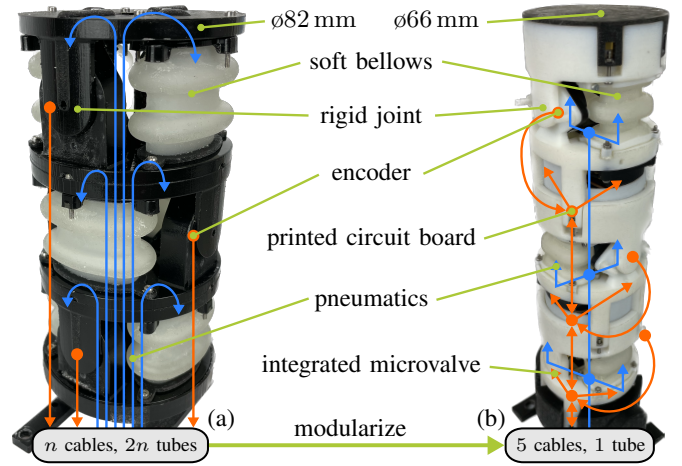}
		\caption{From semi-modular~(a) to modular~(b) ASRs with $n{=}3$ actuators. Blue lines indicate pneumatic connections and orange lines represent electronics. A scalable design is realized via central supply and communication.} \label{fig:cover}
\end{figure}

\IEEEPARstart{S}{nake} robots show much potential in various applications. Their ability to navigate narrow curvilinear paths enables their use for medical procedures~\cite{Orekhov.2018,BurgnerKahrs.2015b}, search and rescue missions, inspection tasks, and many more~\cite{Hopkins.2009}. Soft materials are beneficial for safe human-robot interaction. Two design paradigms exist: soft continuum robots~(SCRs) vs. articulated soft robots~(ASRs)~\cite{GeorgeThuruthel.2018}. The latter comprise a vertebrate-like structure with rigid links and compliant joints. SCRs consist of deformable soft structures along a continuous backbone with infinite degrees of freedom (DoF). Both are often built from a series of \emph{modular} soft actuators. Thereby, modularity should be defined in different terms: \emph{stackability}, \emph{actuation}, and \emph{communication}. Stackable actuators can be connected consecutively, as shown in Fig.~\ref{fig:cover}(a). However, this is insufficient for a \emph{scalable design} since it is impossible to stack \highlightred{a large} number of actuators due to the required space of cables and tubes. Instead, a centralized supply line and signal routing using a data bus is required.

\highlightred{Fig.~\ref{fig:cover} illustrates the transition from a semi-modular robot to a fully modular ASR, which is the core of our work. The advantages of ASRs are the modeling with conventional methods and the simple mounting of encoders while still being intrinsically compliant compared to impedance-controlled robots with necessary high-bandwith feedback. The position feedback simplifies control of the whole robot's shape instead of expensive sensors based on fiber Bragg grating or complex estimation approaches for SCRs. Our ASR design is scalable since the number of cables and tubes led through the whole body is constant and not dependent on the number of actuators $n$. For long, multi-DoF snake robots, gravitational effects must be taken into account when evaluating the stackability. Similar to nature, environmental contacts can be used as support, but the maximum possible actuation force relative to the actuator weight must be considered to resist gravity for some robot segments. Therefore, lightweight pneumatic drives are suitable, as they have a high miniaturization potential compared to electric motors. The latter is used for the \quot{qbmove} actuators to build ASRs, presented in the pioneering work of Della Santina et al.~\cite{DellaSantina.2017b}. By using pneumatics, our modular actuator is $\SI{37}{\%}$ lighter than \quot{qbmove maker pro} (and $\SI{64}{\%}$ lighter than \quot{qbmove advanced}) at a $\SI{22}{\%}$ reduced cross-sectional area. \highlightredsec{Also, pneumatic systems can be used in areas such as MRI scanners, where electric drives are not permitted. The advantages come at the expense of more restrictive design-related limitations of torque and motion range, delays due to tube connections, and the need for proper sealing.}}

\highlightred{Similar to~\cite{DellaSantina.2017b}, our designs are released as open source\footnote{\tt\footnotesize{\url{https://tlhabich.github.io/sponge}}}$^\dagger$ (OS). This }allows others to enter this field by quickly recreating and further developing such robots. Also, comparing research results related to modeling and controlling such nonlinear systems on a real benchmark system is possible. A similar ambition exists in the area of concentric-tube continuum robots~\cite{grassmann2020ctcr}: Due to the lack of a unified robotic platform, there are many individual early-stage prototypes, most of which are used for robot-based evaluations in only a \textit{single} publication. This is mainly due to publication pressure, as validations with real test benches increase the chance of an accepted paper, and simple systems are often sufficient for this purpose. If, on the other hand, a uniform test bench would be freely available, considerable work in building each prototype from scratch would be saved, and the \textit{focus would be shifted towards major research challenges}. Another advantage is the easy reproducibility of published results~\cite{grassmann2020ctcr}. Some progress has already been made in soft robotics, although there is still room for improvement. We address this by presenting SPONGE~-- the \textbf{s}oft \textbf{p}neumatic r\textbf{o}bot with a\textbf{n}ta\textbf{g}onistic b\textbf{e}llows. The remainder of this paper is structured as follows: Section~\ref{relwork} provides related work and outlines our contributions. Based on this, Sec.~\ref{robots} shows our designs. Experimental results with both robots are presented in Sec.~\ref{experiments}, and conclusions are drawn in Sec.~\ref{conclusion}.
%\citep{book}
\section{Related Work}\label{relwork}

Several designs for \highlightred{fluidic soft snake} robots are shown in Table~\ref{tab:overview}. Many ASRs and SCRs are \textit{only} stackable~\highlightred{\cite{Habich.2023,Skorina.52016,Best.2016,Chen.2020,Phillips.2018,Xu.2022,Wang.2021e,Cianchetti.2014,fetch,Jiang.2016,Bruder.2023}}, where each chamber is connected to external valves with one tube each. This hinders miniaturization, as any sensor cables and tubes must be led through the robot body. Thus, the maximum stack length is limited by the robot diameter. Integrating a pneumatic random-access memory~\cite{Hoang.2021} reduces the number of external valves \highlightred{and embedded fluidic control circuits~\cite{Rothemund.2018,Zhai.2023} can even replace integrated solenoid valves}. However, this does not reduce the number of tubes inside the robot. A fully scalable robot design must also be \emph{modular in terms of actuation and communication}. The latter can be achieved by integrating printed circuit boards (PCBs). Signals are then transmitted via a serial bus system with a daisy-chain structure.

\highlightred{Different approaches to modular actuation exist: Pumps can be integrated, resulting in large-scale soft robots~\cite{Li.2022}. Instead, the number of tubes can be reduced via embodied intelligence~\cite{vanRaemdonck.2023}, which only realizes a limited number of input sequences for each actuator. The most common approach is the integration of microvalves, which are connected to a single pressure-supply line for a scalable design.} Some stackable SCRs~\cite{Null.19.09.2022,Onal.2013} are modular in actuation and follow that principle. They all comprise integrated valves connected to a supply line. \highlightred{However, modular communication is not provided since valves and sensors are individually wire-connected to an external controller.}

Six SCRs meet all modularity criteria. Wireless communication is used in the large-scale design consisting of pumps~\cite{Li.2022}. To achieve modular actuation, integrated valves connected to a central pressure supply are utilized in~\cite{Ohno.2000,Robertson.2017,Robertson.2021,Wan.2023,Ikuta.2006}. A communication PCB with a bus system is implemented in~\cite{Ohno.2000,Robertson.2017, Robertson.2021,Wan.2023}. When using custom band-pass valves without integrated sensors~\cite{Ikuta.2006}, PCBs are unnecessary since modular actuation is realized by varying the supply pressure.

The number of different ASR/SCR designs is steadily increasing. They are usually described \textit{briefly} within their respective publications, which complicates reproduction. In one work~\cite{Liljeback.2008}, the first and only fully modular fluidic ASR consisting of integrated valves and a PCB is published. \highlightred{Due to the page limit, the design description takes less than two pages and covers neither the manufacturing/assembly nor their PCB design or any code examples for a test bench.} Further, no part numbers of valves/sensors or CAD files are given. This is not an isolated case, meaning that the \textit{reproducibility of experimental setups is often not possible} due to the brevity of publications. However, all such resources are beneficial for the community. An OS platform, such as~\cite{B.Deutschmann.2022, Grassmann.2024} for tendon-driven continuum robots is desirable.

Efforts in this direction have already been made within the soft-robotics community by publishing an openly accessible test-bench design~\cite{Shi.432023472023} or via the Sorotoki toolkit\highlightred{~\cite{Caasenbrood.2024}}, including a non-modular SCR design. The largest project towards an OS platform is the Soft Robotics Toolkit~\cite{Holland.2014} -- a website where research teams can publish their robot systems. Until now, 17 different actuators with detailed descriptions and 3D models are available. Three SCR designs~\cite{Cianchetti.2014,fetch,Jiang.2016} from Table~\ref{tab:overview} are published on this platform, which are neither modular in terms of actuation nor communication. \highlightred{Recently, Bruder et al.~\cite{Bruder.2023} published their stackable SCR open source, which consists of antagonistic McKibben actuators and has no modular actuation due to external valves.}

\begin{table}[b]
	\centering
	\caption{Modularity within \highlightred{fluidic} soft snake robots}
	\begin{tabular}{|c|c|c|c|c|}
		\hline
		%		\multirow{2}{*}{ref.} & \multirow{2}{*}{type} &  \multicolumn{2}{c|}{modular in terms of ...} & \multirow{2}{*}{OS}\\
		%		\cline{3-4}
		%		& & stacking & actuation \& comm. & \\
		reference& type& stacking & \parbox[c][20pt]{2.1cm}{\highlightred{modular actuation \& communication}} & OS \\
		\hline
		\cite{Habich.2023,Best.2016,Skorina.52016} & ASR & \cmark & \xmark & \xmark \\
		\hline
		\highlightred{\cite{Chen.2020,Xu.2022,Wang.2021e,Null.19.09.2022,Onal.2013,Phillips.2018}}& SCR & \cmark & \xmark & \xmark \\
		\hline
		\highlightred{\cite{Cianchetti.2014,fetch,Jiang.2016,Bruder.2023}} & SCR & \cmark & \xmark & \cmark \\
		\hline
		\cite{Li.2022,Ikuta.2006,Ohno.2000,Robertson.2017,Robertson.2021,Wan.2023} & SCR & \cmark & \cmark & \xmark \\
		\hline
		\cite{Liljeback.2008} & ASR & \cmark & \cmark & \xmark \\
		\hline
		SPONGE & ASR & \cmark & \xmark $\rightarrow$ \cmark & \cmark \\
		\hline
	\end{tabular}
	
	\label{tab:overview}
\end{table}
To summarize, few SCR and ASR designs in the literature take advantage of a fully modular approach. Only one modular fluidic ASR exists with an incomplete description of the system. Our work \textit{contributes} to this field: \textbf{1)}~Building up on our previous actuator design~\cite{Habich.2023}, a semi-modular ASR is presented as a starting point for \textbf{2)}~the transition to a fully modular system with integrated microvalves, central pneumatic supply, and serial communication. \textbf{3)}~Both robots are controlled in experiments to underline their functionality. \textbf{4)}~For long-term operation, the printed soft bellows are replaced by cast ones. This leads to a significantly longer lifetime, as shown in fatigue tests. \textbf{5)}~We publish all materials on an open-source platform, including CAD models, circuit diagram, detailed manufacturing and assembly descriptions, lists of purchased parts \highlightred{and code for test-bench operation}. Our \textit{accessible} robot design, therefore, provides \textit{reproducibility} and \textit{comparability} of research results within modeling and control of such nonlinear systems.
\section{Modular Articulated Soft Robots}\label{robots}
After describing some preliminaries (\ref{preliminaries}), the semi-modular ASR design is introduced (\ref{side_actuator}), and the transition to the fully modular ASR is presented (\ref{main_actuator}). For controlling both robots, the different architectures are explained (\ref{control}). \highlightred{After describing the casting of the bellows for long-term operation (\ref{molded_bellows}), the two robots are compared (\ref{comparison}).}

\subsection{Actuator Preliminaries}\label{preliminaries}
Each actuator $i {\in} \{1,...,n\}$ consists of a one-DoF joint with two antagonistically arranged soft bellows and rigid frames. These are printed using Polyjet technology\footnote{Stratasys Objet350 Connex3, multi-material printing with VeroBlackPlus/VeroWhitePlus (rigid) and \mbox{Agilus30} (flexible)}. Note that in the course of this research, the color of the rigid material was changed from black to white, with the same material properties, due to availability~(cf.~Fig.~\ref{fig:cover}). An ASR is actuated with pressures $\mm{p}{=}[\mm{p}_1\transpose,\mm{p}_2\transpose,...,\mm{p}_n\transpose]\transpose{\in}\R^{2n}$, where each joint angle~$q_{i}$ depends on the difference $\Delta p_i{=}p_{i1}{-}p_{i2}$ of the bellows pressures $\mm{p}_i{=}[p_{i1},p_{i2}]\transpose$. For $q_i{=}0$ (both frames parallel to each other), the bellows are assembled in a slightly compressed configuration. This is advantageous as it lessens the mechanical stress at the maximum joint angle. Since $q_{i}$ only depends on the pressure difference, varying the mean pressure of both bellows changes the joint stiffness~\cite{Habich.2023}.

\subsection{Semi-Modular SPONGE}\label{side_actuator}
\begin{figure}[t]
	\vspace{2.5mm}
	\centering
	\resizebox{1\linewidth}{!}{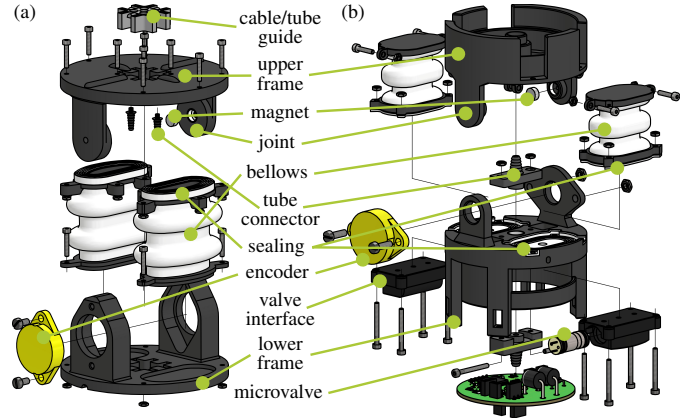}
	\caption{SPONGE designs: (a) The semi-modular actuator \highlightred{(diameter: $\SI{82}{\milli \meter}$, height: $\SI{52}{\milli \meter}$)} consists of soft bellows, a discrete joint, and an encoder. (b) Fully modular design \highlightred{(diameter: $\SI{66}{\milli \meter}$, height: $\SI{94}{\milli \meter}$)} with integrated microvalves and serial communication.} \label{fig:actuators_explosion}
	\vspace{-2mm}
\end{figure}

The actuator shown in Figures~\ref{fig:cover}(a) and~\ref{fig:actuators_explosion}(a) is termed semi-modular since all cables and tubes are led through the whole body to external valves or the controller. It comprises a rigid frame, two soft air bellows, and a Hall encoder \highlightred{(Megatron ETA25K, resolution: $0.09\degree$)}. The latter is mounted to the lower frame opposing the magnet, which is pressed into a soft lip in the joint shaft. The rigid frame is split into an upper and a lower part connected through the one-DoF joint. Shafts can be slightly bent to connect with their corresponding holes in the lower frame, which prevents lateral motion without any further parts. A printed guide in the center of each frame simplifies assembly with many tubes and cables.

Pneumatic channels are printed \textit{within} the upper frame and connect the bellows with the tubes via connectors. The tubes are attached to the connectors by press fit. These miniaturized connectors are separate 3D-printed parts that are press-fitted into the upper frame, as the printing orientation significantly affects their breaking strength. Printing the parts independently saves a considerable amount of support material. Also, the thin connectors for tubes with an outer diameter of $\SI{2}{\mm}$ can be easily replaced in case of breakage instead of reprinting a whole upper frame.

The bellows can be replaced in case of failure due to fatigue effects over time (cf. Sec.~\ref{long_term_val}). It consists of two rigid platforms connected by a soft membrane and two folds. Air can be directed into the bellows through the open upper platform. The latter also ensures that support material can easily be removed from the hollow bellows. A soft, ribbed layer around the upper platform seals the bellows when connected to the upper frame. \highlightred{Screws fasten the upper/lower platform and connect adjacent actuators.} The printed bellows withstands a maximum working pressure of $p_\ind{max}{=}\SI{0.35}{\bar}$. A further increase in pressure expands the joint-angle range but also considerably reduces the lifetime. In this work, the pressure thresholds were tuned iteratively for \textit{all} bellows with experiments to resolve this trade-off.

A previous iteration of this design has been presented in~\cite{Habich.2023} for evaluating a control algorithm. We further developed the design in different ways: \textbf{1)}~The rectangular design was replaced by a circular shape, giving the robot a snake-like appearance with a reduction in diameter by~21\%. \textbf{2)}~Five actuators can still be stacked despite the smaller diameter. This was achieved by miniaturizing the tube connectors for thinner tubes. \textbf{3)}~Through the iterative redesign of the soft bellows, the achievable joint-angle range has been more than doubled to~\SI{37}{\degree}, which results in significantly higher dexterity. \textbf{4)}~\highlightred{The Hall encoder is replaced by a model with four times higher angular resolution, which also better integrates into the snake-robot design due to its flat dimensions.}

\subsection{Modular SPONGE}\label{main_actuator}

The modular concept is shown in Fig.~\ref{fig:actuators_explosion}(b). \highlightred{Several similarities to the semi-modular design exist, such as the split frame, bellows design, joint assembly, screw mountings, and same encoder.} The bellows are also replaceable, but the smaller soft membrane has a reduced thickness and only one fold. This configuration results in a maximum working pressure of $p_\ind{max}{=}\SI{0.3}{\bar}$ and yields a good compromise between bellows stiffness and maximum elongation for this particular actuator design. The lower platform is open and mounted in its respective slot in the lower frame, coated with a soft, ribbed sealing. 

\begin{figure}[t]
	\vspace{2.5mm}
	\centering
	\resizebox{1\linewidth}{!}{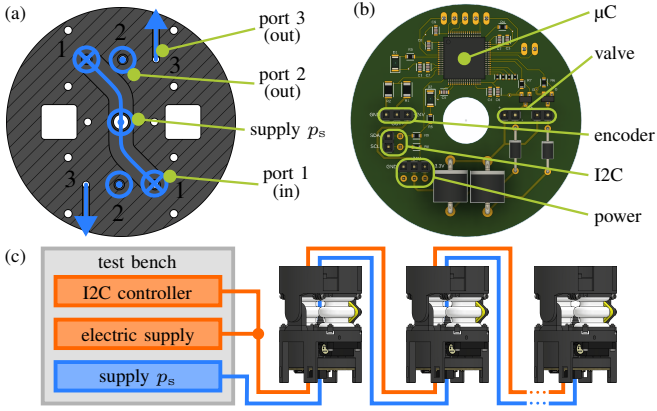}
	\caption{(a) Sectional view with printed pneumatic channels inside the lower frame. Pressure supply $p_\ind{s}$ is directed to port~1 and inflates the bellows (port~2). The bellows is depressurized using port~3. (b) Design of each PCB mounted between adjacent actuators. (c)~Modular test-bench architecture.}
	\label{fig:communication}
	\vspace{-2mm}
\end{figure}

A central pneumatic supply line and serial communication are integrated to modularize the semi-modular ASR. Therefore, the number of cables and tubes led through the body remains constant for ASRs of arbitrary length. For this purpose, two 3/2-way solenoid microvalves \highlightred{(Staiger VA 304-913)} and a PCB are added to each actuator. Both frames contain holes as channels for the electronic cables. A tube connector is located on the upper and lower side of the lower frame for attaching the pneumatic supply line with an outer diameter of $\SI{6}{\mm}$. Tube segments interconnect adjacent actuators and form the central pressure line, closed in the last actuator of an assembled ASR. A printed lid is attached to the last actuator to (optically) close the robot and can be replaced by an end-effector tool.

Printed pneumatic channels \textit{inside} the lower frame are used to distribute the pressurized air to the valves, as illustrated in Fig.~\ref{fig:communication}(a). The supply pressure $p_\ind{s}$ is directed to the normally closed port~1 of each valve. By opening the valve, the air flows from port~1 through port~2 inside the bellows. When the valve is closed, the air inside the bellows is released via port~2 and port~3 to the environment. The valves are plugged into 3D-printed interfaces mounted under the lower frame. Since the valves rely on sealing rings requiring smooth surfaces, the interfaces are printed using stereolithography\footnote{Formlabs Form2, material: BlackV4 (rigid)}. The lower frame is sealed with a soft, ribbed layer to connect the interfaces airtight.

Fig.~\ref{fig:communication}(b) shows the PCB consisting of a microcontroller \highlightred{(\textmu C, STMicroelectronics STM32F401)}, circuits for reading the sensor signal and controlling the valves as well as wire connections for hardware, inter-integrated circuit (I2C) bus lines and power supply. The \textmu C converts the analog sensor signal and controls the valves using MOSFET circuits. Data can be streamed to a computer using a USART connection.

The general structure of communication and supply is illustrated in Fig.~\ref{fig:communication}(c), whereby each actuator is configured as an I2C \highlightred{target}. A development board \highlightred{(STMicroelectronics NUCLEO-F401RE)} serves as the \highlightred{I2C controller}, which can be connected to the test bench (cf.~\ref{test_bench}) via EtherCAT using a shield board \highlightred{(AB\&T EasyCAT)}. The development board serves as an interface between the test bench and the individual PCBs. Only two cables are needed for the I2C bus, regardless of the number of actuators/\highlightred{I2C targets}. However, it is unsuited for long distances between integrated circuits and has a limited communication speed. \highlightred{Note that }in fast mode ($\SI{400}{kbit/s}$ transmission speed), twelve \highlightred{I2C targets} can be controlled with the test bench's sampling frequency of $\SI{1}{\kHz}$. For $n{>}12$ actuators, a different protocol must be used, or the sampling frequency must be reduced. \highlightred{Due to $n{=}3$ in this work, I2C with the selected sampling time is suitable.} Five wires in total are led through a robot \highlightred{with $n{\leq}12$ actuators}, including three wires for the power supply. Similar to the pneumatic supply, cable segments are connected to the lower/upper sides of each PCB.

\subsection{Control Architectures}\label{control}
\begin{figure}[t]
	\vspace{2.5mm}
	\centering
	\resizebox{1\linewidth}{!}{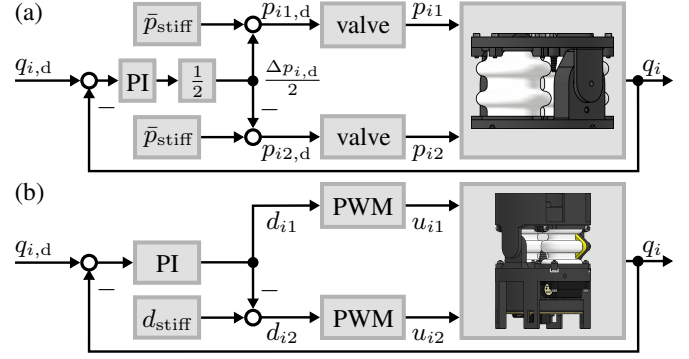}
	\caption{Control architectures: (a) The semi-modular actuator is controlled by adjusting desired bellows pressures, which are set by the integrated pressure controllers of the \textit{external} proportional valves. (b) The modular actuator consists of \textit{integrated} binary valves and is controlled via PWM.}
	\label{fig:control_architectures}
	\vspace{-2mm}
\end{figure}

Proportional-integral (PI) controllers with anti-windup are used to move each actuator to a desired joint angle $q_{i,\ind{d}}$, illustrated in Fig.~\ref{fig:control_architectures}. The joint angle of the semi-modular actuator can be controlled by computing desired pressures $p_{i1,\mathrm{d}}$ and $p_{i2,\mathrm{d}}$. To resolve the infinite number of pressure combinations for each joint angle, the PI controller regulates the pressure around $\bar{p}_\ind{stiff}$. \highlightred{For this purpose, the controller calculates the desired pressure difference $\Delta p_{i,\ind{d}}$ depending on the position error. This quantity is divided in half between the two desired pressures, which results in $p_{i1,\ind{d}}{=}\bar{p}_\ind{stiff}+\Delta p_{i,\mathrm{d}}/2$ and $p_{i2,\ind{d}}{=}\bar{p}_\ind{stiff}-\Delta p_{i,\mathrm{d}}/2$.} External proportional valves with integrated pressure control actuate each bellows individually and set these desired pressures. The mean pressure~$\bar{p}_\ind{stiff}$ could be changed to adjust the joint stiffness, cf.~\cite{Habich.2023}.   

Unlike the proportional valves in the semi-modular robot, the fully modular actuator is equipped with on/off microvalves. Thereby, the air either flows into or out of the bellows. However, continuous bellows pressures can be achieved via pulse-width modulation (PWM) \highlightred{with frequency $f_\ind{PWM}$} of binary valve states $\mm{u}_i{=}[u_{i1},u_{i2}]\transpose$ of the first valve $u_{i1}$ and the second valve $u_{i2}$. The joint angle is controlled by manipulating PWM duty cycles $d_{i1}$ of the first and $d_{i2}{=}d_\ind{stiff}{-}d_{i1}$ of the second valve. Analogous to the semi-modular actuator, the joint stiffness could be varied by adjusting $d_\ind{stiff}$. Since stiffness control is not the scope of this paper, we set $\bar{p}_\ind{stiff}{=}\frac{p_{\mathrm{max}}}{2}$ and $d_\ind{stiff}{=}1$. The respective control architecture is executed in parallel for $n$ actuators.

\subsection{Cast Bellows for Long-Term Operation}\label{molded_bellows}
Printed bellows are problematic for long-term use: The material ages over time and becomes porous. In addition, the layer-by-layer production causes a failure after just a few load cycles (cf. Sec.~\ref{long_term_val}). \highlightred{This problem could be solved for further researched Polyjet printers/materials in the future, but it is currently one of the main drawbacks of this technology, along with the monetary costs.} Silicone casting is \highlightred{a promising} alternative, as the entire membrane cures homogeneously. \highlightred{For both robots, the manufacturing of cast bellows is presented to replace the printed ones.}

\begin{figure}[t]
	\vspace{2.5mm}
	\centering
	\resizebox{1\linewidth}{!}{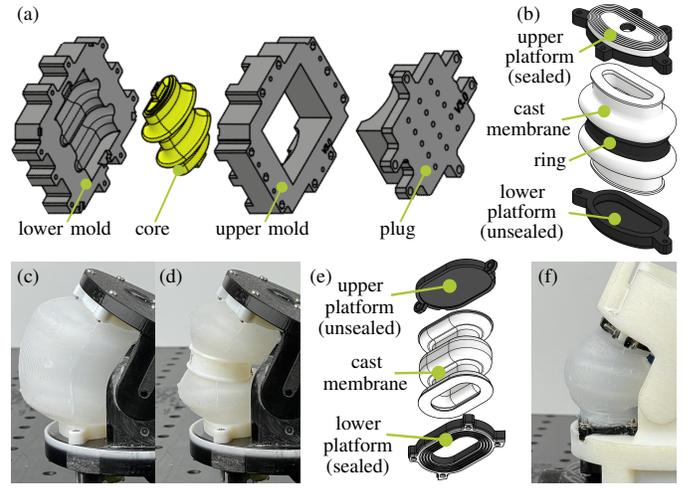}
	\caption{\highlightred{(a) Molds for casting the bellows and (b) post-casting assembly for semi-modular design, ballooning of the cast bellows~(c) without and (d)~with a ring at \highlightred{$p_\ind{max}{=}\SI{0.5}{bar}$} (semi-modular), (e) post-casting assembly and (f) expanded bellows at $p_\ind{max}{=}\SI{0.3}{bar}$ (modular)}}
	\label{fig:molding}
	\vspace{-2mm}
\end{figure}

To cast the bellows membrane, the parts from Fig.~\ref{fig:molding}(a) are 3D-printed\footnote{Ultimaker 3, material: PLA filament}. Lower mold, core \highlightred{and} upper mold are assembled and closed with a plug after filling in the silicone \highlightred{(Smooth-On Dragon Skin 15)}. A vacuum oven is used to remove any trapped air bubbles. The cured membrane of the semi-modular design is equipped with a printed ring and glued to the printed upper/lower platforms, which is visualized in Fig.~\ref{fig:molding}(b). The assembled cast bellows has an identical geometry to its printed version, except for the ring and higher platform edges to increase the adhesive surface. The ring prevents high deformation, which is visualized in Fig.~\ref{fig:molding}\mbox{(c)--(d)}. This ballooning effect is undesirable, as the robot's diameter significantly increases, and the hysteresis effects are considerably higher. The ballooning effect is significantly less with printed bellows, so no radial stiffening is necessary. \highlightred{The cast bellows withstands a higher pressure of $p_\ind{max}{=}\SI{0.5}{bar}$ due to the different material combined with weak ballooning by the ring.} 
	
\highlightred{Similar molds with different dimensions are used for the bellows of the modular design. The post-casting assembly and the expanded bellows at $p_\ind{max}{=}\SI{0.3}{bar}$ is illustrated in Fig.~\ref{fig:molding}(e)--(f). Due to the different design with smaller dimensions, less ballooning occurs. A ring is, therefore, not necessary. This could be added in future iterations to further increase $p_\ind{max}$. A redesign of the bellows would also be necessary, as there is currently only one fold due to its size.}

\highlightred{\subsection{Semi-Modular vs. Modular}\label{comparison}
Finally, the main differences between the ASRs are discussed.
\subsubsection{Dimensions} Compared to the semi-modular design (diameter: $\SI{82}{\mm}$, height: $\SI{52}{\mm}$, weight: $\SI{0.150}{\kilogram}$), a smaller diameter of $\SI{66}{\mm}$ is possible due to a single pressure line, modular communication and smaller bellows. This is at the expense of an increased height ($\SI{94}{\mm}$) and weight ($\SI{0.163}{\kilogram}$) due to microvalves and a PCB.
\subsubsection{Modularity} The number of cables and tubes for the semi-modular robot scales with $n$, whereas this is constant for the modular system. Thus, the semi-modular design is only stackable (given the chosen dimensions) for five actuators and the modular system \textit{theoretically} for an unlimited number. In practice, this is also limited by flow resistance in the supply line, data-bus constraints, and the maximum possible torques to resist gravity. The latter can be maximized by using stiffer bellows that allow higher pressures.
\subsubsection{Pneumatics} In the semi-modular robot, the bellows pressures can be precisely set and measured due to the proportional valves. This makes it more practical for research, e.g., on (learning-based) system identification. Due to the thinner tubes, significantly lower volume flows are possible than in the modular robot, which leads to delays. Instead, the binary microvalves have the disadvantage of generating noise due to the PWM and a limited lifetime. The PWM frequency is set to $f_\ind{PWM}{=}\SI{100}{\hertz}$. According to the manufacturer, the microvalve was successfully tested for 500~million switching cycles. Some samples already show internal leakages after 100~million switching cycles. Therefore, the valve could be damaged due to mechanical wear after approx. $\SI{278}{\hour}$--$\SI{1389}{\hour}$ of operation. Regarding pressure measurement in the modular robot, additional sensors could be integrated, requiring more space.
\subsubsection{Handling} Due to the compact design, the assembly of the modular robot with a central pressure line and PCBs is more complex. In areas where magnetic components are not permitted, such as MRI scanners, the modular robot cannot be used due to the solenoid valves. After replacing the Hall encoder with an MRI-safe alternative and using suitable screws/nuts, the semi-modular design can be used there.}

\highlightred{In summary, the semi-modular robot is suitable as a research platform for modeling, identification, and control due to the precisely adjustable (and measured) pressures as well as the simpler assembly. In contrast, due to its scalability, the modular robot forms the basis for multi-DoF snake robots. Hybrid semi-modular designs are also conceivable so that, e.g., encoders are read out with I2C, which would double the maximum number of stackable actuators to $n{=}10$.}
\section{Experiments}\label{experiments}
 After describing the test bench (\ref{test_bench}) and investigating the airtightness (\ref{supply}), the functionality of both robots is demonstrated in experiments (\ref{stacked_act}). Further, long-term operation with the different bellows is investigated (\ref{long_term_val}). \highlightred{During all experiments, the maximum pressure $p_\ind{max}$ for each bellows type is limited to the tuned thresholds of Sec.~\ref{robots}.}

\subsection{Test Bench}\label{test_bench}
\begin{figure}[t]
	\vspace{2.5mm}
	\centering
	\resizebox{1\linewidth}{!}{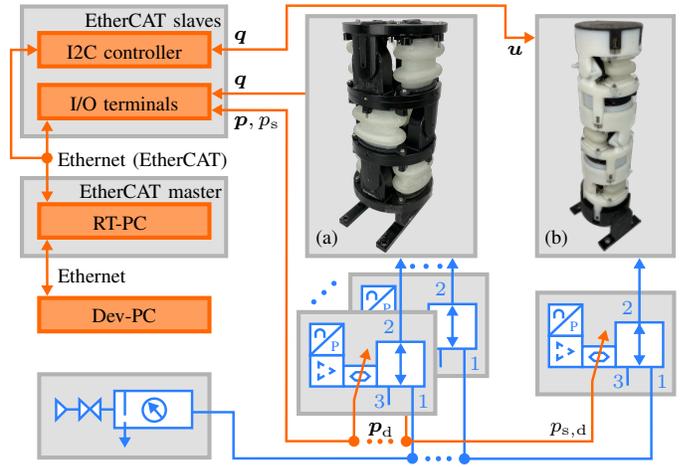}
	\caption{Test-bench architecture: Orange lines indicate electronic signals. For the sake of clarity, one single arrow illustrates the desired pressures $\mm{p}_\ind{d}$ for semi-modular SPONGE~(a), the desired supply pressure $p_\ind{s,d}$ for modular SPONGE~(b), and all corresponding measured pressures $\mm{p}$, $p_\ind{s}$. Blue lines indicate pneumatics. The semi-modular robot is connected directly to EtherCAT I/O terminals. The modular robot is linked via an \highlightred{I2C controller}.} 
	\label{fig:testbench}
	\vspace{-2mm}
\end{figure}

The test bench is equipped with a pneumatic supply, a development computer (Dev-PC), a real-time computer (RT-PC), EtherCAT slaves, as well as several proportional piezo valves \highlightred{(Festo VEAA-B-3-D2-F-V1-1R1, resolution: \SI{5}{\milli\bar})} with integrated pressure control. Its architecture is depicted in Fig.~\ref{fig:testbench}. An array of piezo valves is connected in series with the supply unit. \highlightred{The latter could be extended by an additional reservoir to compensate for supply-pressure fluctuations, which are negligible in our laboratory}. The semi-modular ASR requires $2n$ valves. For the modular robot, one valve must be connected to control the supply pressure $p_{\mathrm{s}}$. Data acquisition and real-time system control with a cycle time of $\SI{1}{\milli\second}$ is possible. The controller is designed and compiled on the Dev-PC using Matlab/Simulink. The RT-PC runs the compiled model. Data can be streamed for visualization/logging, and settings can be altered during runtime. The communication is realized using EtherCAT protocol and the corresponding open-source tool EtherLab with an added external-mode patch\footnote{\tt{\url{https://github.com/SchapplM/etherlab-examples}}}. 

Sensor data (pressures $\mm{p}$, supply pressure $p_\ind{s}$, joint angles~$\mm{q}$ of the semi-modular robot) and desired pressures~$\mm{p}_\mathrm{d}$~(semi-modular) as well as the desired supply pressure~$p_\ind{s,d}$~(modular) are read or set over EtherCAT terminals \highlightred{(Beckhoff EL3702, EL4102)}. The modular ASR is connected to the \highlightred{I2C controller}, an independent EtherCAT slave with inputs and outputs. Joint angles~$\mm{q}$ of the modular robot and (binary) valve configurations $\mm{u}{=}[\mm{u}_1\transpose,\mm{u}_2\transpose,...,\mm{u}_n\transpose]\transpose{\in}\R^{2n}$ are communicated via I2C.

\subsection{Ensuring Airtightness}\label{supply}
Using pneumatic actuation and 3D printing necessitates focusing on the systems' airtightness. In the semi-modular robot, the bellows are directly connected to the external valves, and air tightness is achieved by printed sealing lips on the bellows. Due to the higher complexity of the modular ASR with a central pressure line, some design iterations with subsequent leak testing are required. Since numerous actuators share the supply line, and the PWM control leads to pressure fluctuations, a high supply pressure $p_\ind{s}$ is needed. To build \highlightred{\textit{long}} snake robots, a high pressure supply is also required to provide sufficiently high torques to resist gravity. The leakages are investigated by gradually increasing the desired supply pressure $p_\ind{s,d}$ to determine the maximum possible supply pressure $p_\ind{s,max}$. All microvalves are closed during the experiment so that the bellows are decoupled from the supply line. Therefore, any pressure losses are solely linked to the other printed parts. 
\begin{figure}[t]
	%\vspace{2.5mm}
	\centering{\includegraphics[width=\linewidth]{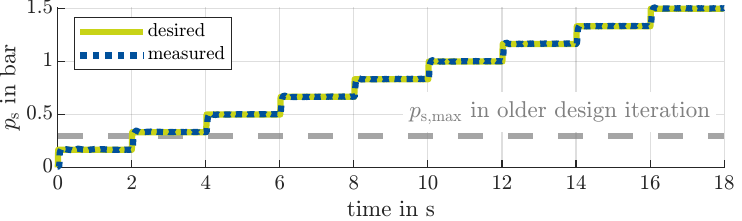}}
	\caption{Airtightness of modular SPONGE: The maximum supply pressure in an older design iteration was limited to $p_\ind{s,max}{\approx}\SI{0.3}{bar}$, which hinders the stackability required for long snake robots.}
	\label{fig:supply_pressures}
\end{figure}

With an \textit{early-stage design iteration}, this experiment demonstrates a limited supply pressure $p_\ind{s,max}{\approx}\SI{0.3}{bar}$. The main reason \highlightred{\textit{was}} splitting the lower frame into two parts so that the support material in the narrow pneumatic channels could be easily removed. \highlightred{Despite sealing, small leakages between the split lower frame were present}. Although this pressure would be sufficient for our test configuration with $n{=}3$ actuators and maximum bellows pressure $p_\ind{max}{=}\SI{0.3}{bar}$, with each additional actuator, the supply pressure would further decrease due to additional leakages. Thus, the system \textit{would} have limited stackability. To address this, the lower frame was redesigned as \textit{one} part with integrated pneumatic channels, shown in Fig.~\ref{fig:communication}(a). For this final design iteration, Fig.~\ref{fig:supply_pressures} plots the measured and desired pressure values during the leak experiment. The test was performed up to a desired pressure of $p_\ind{s,d}{=}\SI{1.5}{bar}$, which could be maintained in the central line. Accordingly, the leakages were eliminated, which is the basic requirement to build a long snake robot. \highlightred{Note that an adjustment of the soft bellows would also be required to lift the weight of such a robot with, e.g., $n{=}20$ actuators. This can be achieved by using a stiffer material for the bellows membrane to make high pressures/torques possible. The modular system allows the bellows to be easily replaced, which is already demonstrated in this work by the cast bellows.} \highlightredsec{Footage of experiments investigating the ability to resist gravity with $n{=}3$ can be seen in the supplementary video of this article\footnote{\highlightredsec{\tt\url{https://youtube.com/watch?v=TMLpRXZHuLA}}}.}

%\subsection{Position Control of Single Actuators}\label{single_act}
%
%%\begin{figure}[t]
%
%%	\centering{\includegraphics[width=\linewidth]{traj_single_act.pdf}}
%%	\caption{Response of the single actuators to a series of angular jumps. (a): semi-modular actuator. (b): fully modular actuator.}
%%	\label{fig:traj_single}
%%
%%\end{figure}
%To obtain the movement behaviors, a trajectory containing a series of angular jumps is specified. Fig.~\ref{fig:traj_single} shows how both actuators follow their trajectory. The differences in bellows designs and frame proportions lead to different angular ranges and joint stiffnesses. To account for that, the trajectories are adjusted according to the actuators' angular ranges. Here, they are $q_{\mathrm{max}}\approx\pm20\degree$ for the semi-modular ASR and $q_{\mathrm{max}}\approx\pm30\degree$ for the fully modular one. Both actuators can follow their trajectories well despite the simple control schemes. The semi-modular actuator shows an anisotropy. This can be due to an undetected hole in one bellow or a printing error for example within one pneumatic channel. The modular actuator seems to have a shorter delay. This may be rooted in design characteristics. For both actuators the rise times increase when approaching $q_{\mathrm{max}}$.

\subsection{Position Control}\label{stacked_act}
\begin{figure}[t]
	\vspace{2.5mm}
	\centering{\includegraphics[width=\linewidth]{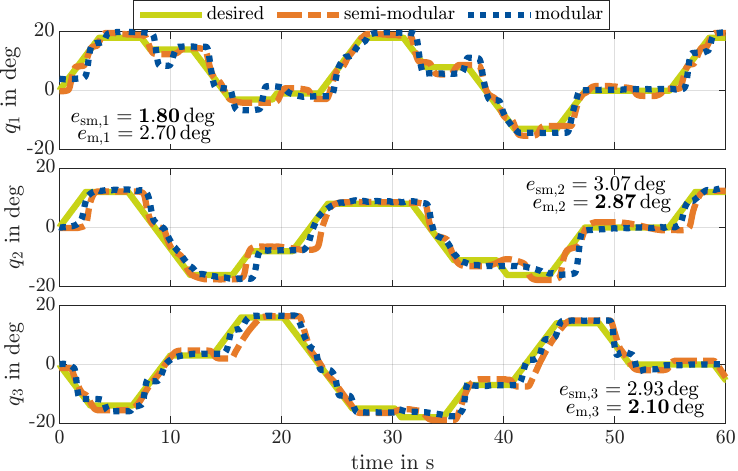}}
	\caption{Control experiments with semi-modular and modular robot}
	\label{fig:traj_stack}
	\vspace{-2mm}
\end{figure}
The functionality of both robots is evaluated using the described control architectures given a ramp trajectory of desired joint angles $\mm{q}_\ind{d}$. The gains of the PI controllers were tuned manually. Fig.~\ref{fig:traj_stack} shows the results, including the root-mean-square error~$e_{\ind{sm,}i}$ of the semi-modular and~$e_{\ind{m,}i}$ of the modular robot comparing the desired and measured joint-angle trajectory for joint $i$. The modular actuator is significantly thinner ($\SI{20}{\%}$ reduced diameter) and longer ($\SI{81}{\%}$ increased height) compared to the semi-modular design. This especially complicates the position control of the first joint and makes it a challenging balancing task against gravity, which can be seen in the slight oscillations. The modular design has a slightly lower position error for the last two joints. This may be related to the fact that with each additional actuator, the length of the thin tubes between bellows and proportional valves increases in the semi-modular design, resulting in a larger dead time in the system. This problem does not occur with the modular SPONGE because the microvalves are mounted directly under the bellows. \highlightred{Another reason may be the lower possible volume flow due to the thinner tubes in the semi-modular robot.} However, both robots exhibit precise trajectory tracking with mean errors of $\bar{e}_\ind{sm}=\SI{2.60}{deg}$ and $\bar{e}_\ind{m}=\SI{2.56}{deg}$.

\subsection{Long-Term Operation}\label{long_term_val}
\highlightred{The suitability for long-term use of all bellows types is investigated by applying $p_\ind{max}$ for $\SI{10}{\second}$ and subsequent venting for $\SI{10}{\second}$. This is repeated to determine the lifetimes. Table~\ref{tab:long_term} shows the results. The printed bellows of both designs already tear within a few minutes, and they are then no longer able to operate the actuator due to leakage. In contrast, the cast bellows with ring (semi-modular) failed after $\SI{72}{\hour}$ and $\mathbf{\SI{36}{\minute}}$ with $\SI{13068}{}$ cycles. The cast bellows without ring (modular) resisted even longer the periodic load and is still usable after ${}\SI{175}{\hour}$ ($\SI{31500}{}$ cycles). This further increase in lifetime can be explained by the fact that no additional ring was installed, and therefore $p_\ind{max}$ was not increased further. In general, the experiments underline the practical applicability of our robots with cast bellows.}

\highlightred{One final remark must be made: Since the bellows' lifetime is significantly increased by casting, all Polyjet parts are non-wearing parts, except the printed parts of the bellows (lower/upper platform and ring). The Polyjet printing used in this work is expensive and is not standard laboratory equipment. Replacing this technology entirely is problematic with the current design, as the soft material is printed directly onto rigid parts for sealing, or to attach the magnet of each encoder. However, laboratories could order all robot parts \textit{once} online, along with some spare parts for the sealed platform. If a bellows ever fails, the ring and unsealed platform of each bellows can be reprinted using conventional technologies, e.g., FDM, and the membrane can be cast.}
\begin{table}[t]
	\vspace{2.5mm}
	\caption{Long-term test \highlightred{(sm${=}$semi-modular, m${=}$modular)}}
	\vspace{-4mm}
	\label{tab:long_term}
	\begin{center}
		\begin{tabular}{|c|c|c|c|}
			%\rule{0pt}{0ex}
			\hline
			\highlightred{bellows type}&$p_\ind{max}$&lifetime\\
			\hline
			\highlightred{sm, 3D-printed}&$\SI{0.35}{bar}$&$\SI{7}{\minute}$ and $\SI{54}{\second}$\\
%			\hline
			\highlightred{sm, cast}&\highlightred{$\SI{0.5}{bar}$}&\highlightred{$\SI{72}{\hour}$ and $\mathbf{\SI{36}{\minute}}$} \\
			\hline
			\highlightred{m, 3D-printed}&\highlightred{$\SI{0.3}{bar}$}&\highlightred{$\SI{15}{\minute}$ and $\SI{2}{\second}$} \\
%			\hline
			\highlightred{m, cast}&\highlightred{$\SI{0.3}{bar}$}&\highlightred{$>\SI{175}{\hour}$} \\
			\hline
		\end{tabular}
	\end{center}
	\vspace{-2.5mm}	
\end{table}
\section{Conclusions}\label{conclusion}
The number of soft-robot designs is steadily increasing. Often, these are only available via brief descriptions in publications that are typically severely shortened. This complicates \textit{accessibility} of the existing designs for reproduction and \textit{reproducibility} of achieved results. In addition, widely varying robot designs hinder \textit{comparability} of research findings. Our open-source soft-robot designs address this, allowing researchers to tackle the major challenges of designing, modeling, and controlling these promising systems. We focus on system modularity, which is especially important when building multi-DoF robots for real-world tasks, such as (redundant) snake robots with many similar actuators.

Starting with a stackable actuator, the necessary adaptations to achieve a fully modular robot are described. This is mainly realized by integrating microvalves, a central supply line, and circuit boards. A leak experiment demonstrates the airtightness of the modular system. Control architectures are implemented to move the two robots along desired (randomly chosen) joint-angle trajectories. Position tracking is done using manually tuned controller gains and without a complex control approach. This underlines the functionality of both systems. The lifetime of the bellows is examined with a fatigue test, whereby the presented cast versions resist significantly longer than printed ones. By open-source publication of all soft- and hardware files, anyone can easily replicate and further develop SPONGE in their lab. We propose using the semi-modular system as a research platform for robot-based evaluations, e.g., of control algorithms. Instead, the modular robot is the starting point to build multi-DoF snake robots due to its scalable design.

In future work, the maximum working pressure of the bellows could be increased via different materials or design optimization of the bellows via finite-element modeling. This would allow higher torques, enabling long snake robots to be moved against gravity. \highlightred{Redesigning the actuator could also further increase the achievable joint-angle range and, therefore, the robot's dexterity.} Furthermore, learning-based control approaches~\cite{Laschi.2023b} enable more precise position tracking of the nonlinear system. This becomes especially relevant for robots with many DoFs and more dynamic motion when simple PI controllers fail and conventional modeling approaches within model-based control are not sufficiently accurate due to the viscoelastic bellows or the friction.

%\addtolength{\textheight}{-8.7cm}   % This command serves to balance the column lengths

%%%%%%%%%%%%%%%%%%%%%%%%%%%%%%%%%%%%%%%%%%%%%%%%%%%%%%%%%%%%%%%%%%%%%%%%%%%%%%%%

%%%%%%%%%%%%%%%%%%%%%%%%%%%%%%%%%%%%%%%%%%%%%%%%%%%%%%%%%%%%%%%%%%%%%%%%%%%%%%%%
\section*{Author Contributions}
TLH, MB created the semi-modular robot. JH, TLH developed the modular robot. DL supported the PCB design. TLH, JH, MB carried out the experiments. TLH guided the manuscript writing including revision process. All authors contributed to the manuscript. MS supervised the research.

%%%%%%%%%%%%%%%%%%%%%%%%%%%%%%%%%%%%%%%%%%%%%%%%%%%%%%%%%%%%%%%%%%%%%%%%%%%%%%%%

\section*{Acknowledgment}
We thank Sarah Kleinjohann and Jan Christoph Haupt for contributing to the semi-modular design and Dennis Bank for the PCB manufacturing.

%%%%%%%%%%%%%%%%%%%%%%%%%%%%%%%%%%%%%%%%%%%%%%%%%%%%%%%%%%%%%%%%%%%%%%%%%%%%%%%%
\bibliographystyle{IEEEtran}
\bibliography{literatur}

\end{document}